\documentclass[letterpaper]{article} % DO NOT CHANGE THIS
\usepackage[preprint]{aaai2027}
\usepackage[hyphens]{url}  % DO NOT CHANGE THIS
\usepackage{graphicx} % DO NOT CHANGE THIS
\usepackage{natbib}  % DO NOT CHANGE THIS AND DO NOT ADD ANY OPTIONS TO IT
\usepackage{caption} % DO NOT CHANGE THIS AND DO NOT ADD ANY OPTIONS TO IT
\usepackage{algorithm}
\usepackage{algorithmic}
\usepackage{amssymb}
\usepackage{amsmath}
\usepackage{multirow}

\usepackage{newfloat}
\usepackage{listings}
\DeclareCaptionStyle{ruled}{labelfont=normalfont,labelsep=colon,strut=off} % DO NOT CHANGE THIS
\floatstyle{ruled}
\newfloat{listing}{tb}{lst}{}
\floatname{listing}{Listing}

\usepackage{booktabs}

\title{ReG-SAM: Reference Graph-Driven SAM for 2D Foundational Vessel Segmentation}
\author {
    Donghang Lyu\textsuperscript{\rm 1}\equalcontrib,
    Zichen Zhang\textsuperscript{\rm 1}\equalcontrib,
    Oleh Dzyubachyk\textsuperscript{\rm 1},
    Marius Staring\textsuperscript{\rm 1}
}
\affiliations {
    \textsuperscript{\rm 1} Division of Image Processing, Department of Radiology, Leiden University Medical
Center, Leiden, the Netherlands\\
}

\begin{document}

\maketitle

\begin{abstract}
Vessel segmentation in medical images is essential for many clinical tasks, ranging from diagnosis to treatment planning. However, it remains challenging due to complex vascular morphology and diverse imaging conditions. Existing deep learning methods rarely aim at building a generalizable vessel segmentor across anatomies and modalities. While the Segment Anything Model (SAM) has shown promise for medical image segmentation, its original design does not fully exploit vascular morphology and struggles with fine-grained vascular structures, leading to suboptimal performance. In this paper, we propose ReG-SAM, a SAM-based framework tailored to 2D vessel segmentation that leverages reference graph set for enhancing vascular representations. Specifically, we introduce two modality-aware representations derived from the reference masks: graph prompt embeddings (GPEs) that encode global spatial features from graphs, and vascular prototype embeddings (VPEs) that capture fine-grained modality-specific vessel characteristics from multi-scale feature maps and vascular masks. Since both require vascular masks that are unavailable during inference and require robust modality-aware vascular feature representations, we construct a modality-wise vascular database and develop two reference graph-guided representation learning schemes for estimating GPEs and VPEs using samples from the database rather than ground-truth masks. Extensive experiments across 19 datasets demonstrate that ReG-SAM consistently outperforms existing baselines, even those using manual prompts, particularly on challenging thin vessels. The code and trained models will be made publicly available.
\end{abstract}

% Uncomment the following to link to your code, datasets, an extended version or similar.
% You must keep this block between (not within) the abstract and the main body of the paper.
% Make sure that you do not de-anonymize yourself with these links.
% \begin{links}
%     \link{Code}{https://aaai.org/example/code}
%     \link{Datasets}{https://aaai.org/example/datasets}
%     \link{Extended version}{https://aaai.org/example/extended-version}
% \end{links}

\section{Introduction}

Vessel segmentation, which extracts vascular structures from medical images, plays an important role in clinical diagnosis and treatment planning~\cite{kock2024suitability,brugnara2023deep}. It is considerably more challenging than organ segmentation due to the thin, elongated, and highly branching vascular morphology, large diameter variations, low contrast against surrounding tissues, and frequent discontinuities caused by imaging artifacts. Furthermore, vascular appearance varies substantially across anatomies and imaging modalities, making robust vessel segmentation particularly difficult. Traditional methods are essentially image filters~\cite{frangi1998multiscale} that rely on prior knowledge of vascular morphology. These modality-dependent approaches, although effective in specific scenarios,  often struggle to generalize across diverse clinical settings. With the advent of deep learning, neural networks have demonstrated strong capability in modeling complex vascular patterns, leading to substantial performance improvements. Representative applications include fundus imaging~\cite{wu2019vessel,wang2020hard,ma2024improved,luo2025pa}, X-ray angiography~\cite{iyer2021angionet,zhu2021coronary,he2025deep}, and optical coherence tomography angiography~\cite{liu2022disentangled,li2025direction}. However, most are designed for individual datasets or modalities. Although foundation medical segmentation models have attracted increasing attention, they are not specifically tailored for vessels. Only few foundational methods~\cite{song2025optimized} have been proposed for 2D vessel segmentation across anatomies and modalities, leaving this area largely underexplored. In addition, these methods fail to capture discriminative vascular appearance features and model modality-specific vascular structures, such as distinct branching patterns.

The Segment Anything Model (SAM)~\cite{kirillov2023segment}, a foundation model originally proposed for natural image segmentation, has been successfully adapted to medical domains~\cite{cheng2023sam,ma2024segment}, making it a promising backbone for a general vessel segmentor. In SAM, the prompt encoder is an important component that provides high-level guidance to the mask decoder. However, its prompt types, such as points and bounding boxes, are inherently suboptimal for vessel segmentation. Unlike compact organs, vessels are elongated and highly branched tree-like structures, making point-based interaction unreliable, while randomly selected points may even degrade performance. Bounding boxes are also ineffective, as vessels often span large regions of the image, often covering nearly the entire field of view. In contrast, graphs naturally represent vascular morphology, making them a natural choice for  vessel segmentation prompts. Besides, graph-derived prompt embeddings can include both vascular structural information and modality-specific characteristics, providing effective guidance across anatomies. Another limitation lies in the mask decoder, which was originally designed for generic object segmentation and hence struggles to preserve fine vascular details such as thin branches and vessel bifurcations. Substantial\- appearance variations across imaging modalities further exacerbate this problem, as the decoder block lacks explicit modality-aware vascular representations to guide fine-grained vessel recovery.

In this paper, we present \textbf{ReG-SAM}, a fully-automatic foundation model for 2D vessel segmentation built upon SAM and guided by modality-aware reference graph set. In particular, we introduce two modality-aware representations to enhance the prompt encoder and mask decoder. (1) A graph constructed from the vascular mask of an image is encoded into a graph prompt embedding (GPE), capturing modality-aware vascular morphology for guiding mask decoding. (2) An auxiliary convolutional neural network (CNN) is employed for extracting multi-scale image features that are used for computing a vascular prototype embedding (VPE) along with the vascular mask, yielding modality-specific vascular representations for fine-grained decoding.

However, GPEs and VPEs face two key challenges. The first is their dependence on ground-truth vascular masks that are unavailable during inference. In particular, GPE construction requires graphs derived from ground-truth masks. Existing alternatives construct graphs by partitioning images into patches and treating each patch as a node~\cite{li2026learning,jalali2024vga}, or by generating coarse masks using auxiliary modules~\cite{mishra2021vtg,shin2019deep}. However, such graphs often fail to faithfully represent global vascular structures and generalize poorly across anatomies. Second, the learned representations should accurately capture modality-aware spatial and semantic information while remaining robust to variations in anatomy and imaging modality.

To address these challenges, we propose two reference graph-guided representation learning schemes for GPEs and VPEs. (3) We build a modality-aware vascular database containing mask-derived graphs. During training, reference graphs are randomly sampled from the database and encoded into embeddings, which are encouraged to align with the embedding representation of a query image through dedicated loss functions. Consequently, robust GPEs and VPEs with corresponding modality priors can be estimated from the database during inference without ground-truth masks. (4) We establish a comprehensive benchmark comprising 19~public datasets across anatomies and modalities. Extensive experiments demonstrate that ReG-SAM outperforms existing SAM-based foundational methods for 2D vessel segmentation, even those augmented with manual prompts.

\section{Related Work}
\subsection{Domain-Specific Vessel Segmentation}
Early methods~\cite{frangi1998multiscale,staal2004ridge} rely on handcrafted vascular priors, such as matched filtering, ridge detection, and Hessian-based vessel filtering to enhance curvilinear structures, but require careful parameter tuning and generalize poorly across anatomies and imaging modalities. Although deep learning has substantially improved vessel segmentation performance, most existing methods~\cite{wu2019vessel,wang2020hard,ma2024improved,luo2025pa,iyer2021angionet,zhu2021coronary,he2025deep,liu2022disentangled,li2025direction}  remain domain-specific, being trained and evaluated on individual datasets or modalities. To address the sparsity of vascular annotations in the area, some approaches focus on data synthesis~\cite{ma2021self,kim2022diffusion,lin2023yolocurvseg,zhang2024self,kreitner2024synthetic,zhang2024xa,liang2026self,hao2026dosta}, scaling up training data with synthesized ones, for achieving better segmentation results. Some works incorporate graph information either by constructing graphs from intermediate segmentation masks~\cite{shin2019deep,mishra2021vtg} or by building graph structures directly within feature representations~\cite{xu2024g2vit,ahmed2026tffm} or from the input image via patch-based partitioning~\cite{jalali2024vga,li2026learning}.

\subsection{Foundational Medical Image Segmentation}
Foundational medical image segmentation methods typically build upon SAM and adapt it to medical domains through fine-tuning. SAM-Med2D~\cite{cheng2023sam} and MedSAM \cite{ma2024segment} are two pioneering medical image segmentors in this direction. Subsequent works have focused on improving specific aspects, such as parameter-efficient adaptation using adapters~\cite{wu2025medical,tejero2025sam} and enhancing computational efficiency~\cite{ma2024efficient,lyu2024mcp}. Despite recent advances, foundational 2D vessel segmentation remains largely underexplored. OVS-Net~\cite{song2025optimized} represents the current state-of-the-art general vessel segmentor, employing a hybrid encoder that combines a SAM backbone with a convolutional branch. However, it remains a semi-automatic framework that relies on conventional point prompts and lacks dedicated modeling of vascular morphology.

\section{Methodology}
\begin{figure*}[!t]
\centering
\includegraphics[width=.97\textwidth]{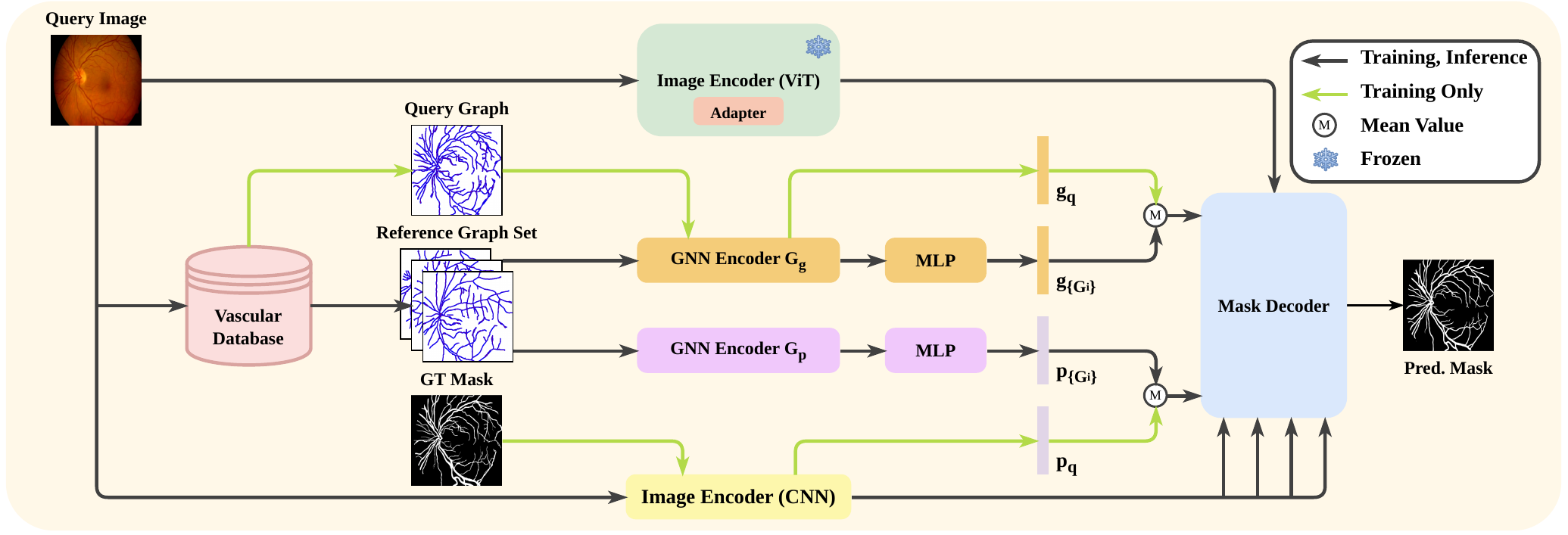}
\caption{An overview of the proposed ReG-SAM. Given a query image, reference graphs are randomly sampled from a vascular database, and their combined embeddings are aligned with the embedding representations of the query image during training.}
\label{fig1}
\end{figure*}

In this section, we present the ReG-SAM for foundational 2D vessel segmentation, as illustrated in Figure~\ref{fig1}. Graph prompt embeddings (GPEs) and vascular prototype embeddings (VPEs) are employed to enhance the segmentation performance. To eliminate their need for ground-truth masks during inference and ensure robust representation learning, we construct a modality-organized vascular database and propose two reference-guided representation learning schemes. Given a query image, $k$ reference graphs are randomly sampled from the database based on the modality of the query image. They are encoded into combined embeddings for aligning with the embedding representations of the query image. During inference, the GPE and VPE are estimated directly from the database without relying on annotated masks, enabling fully automatic and enhanced vessel segmentation.

\subsection{Graph Set Sampling from Vascular Database}
Annotated vascular images are much scarcer than other medical images~\cite{ma2024segment}, making effective use of available data crucial. Moreover, images from the same modality share consistent semantic and imaging characteristics, allowing segmentation to benefit from same-modality references. We therefore construct a modality-organized vascular database containing graphs and their modality labels. A vascular graph $G_{i}=\{V_{i}, E_{i}\}$ is constructed from the mask $\mathbf{M}_{i} \in \{ 0,1 \}^{h\times w}$, where $V_{i}$ and $E_{i}$ denote the node and edge sets, respectively, and $h$ and $w$ are spatial dimensions. Specifically, the mask is first skeletonized~\cite{zhang1984fast,soille1999morphological,lee1994building} and pruned to remove short, noisy branches. Skeleton pixels are then treated as nodes, local neighborhood connections as edges, and pixel coordinates as node features, preserving vessel geometry and topology for graph-based representation learning. Although same-modality images share common imaging characteristics, their anatomical details may vary substantially. Thus, rather than selecting only the most similar samples, which may reduce diversity and limit generalization, we randomly sample references to learn robuster modality-level priors. Given a query image $\mathbf{I}_{q} \in \mathbb{R}^{3 \times h \times w}$ with modality $m_{q}$, we define the corresponding modality-specific candidate set as $S_{m_{q}}$. This set contains the indices of the graphs with the same modality $m_{q}$ except for the graph of the query image itself. Consequently, we randomly sample $k$ indices from $S_{m_{q}}$:
\begin{equation}
N_{q} = \operatorname{RANDOM}_{k}(S_{m_{q}}).
\end{equation}
Finally, the reference graph set $\{\,G_{i} \mid i\in N_{q}\,\}$ is constructed for the query image.

\subsection{Reference-Guided Graph-Prompt Learning}
Since encoding of vascular graphs into embeddings can provide complementary structural information for vessel segmentation, we introduce graph prompt embeddings (GPEs) for capturing modality-aware vascular morphology. As illustrated in Figure~\ref{fig1}, given a query image, its graph is encoded into an embedding to aid the mask decoder, similarly to the sparse prompt embedding in the original SAM. However, the graphs rely on annotated vascular masks that are unavailable during inference. Therefore, we propose a reference-guided graph-prompt learning (RGPL) scheme. We employ a shared graph encoder $\mathcal{G}_{g}$ to encode both the graph $G_{q}$ of the query image and the sampled graph set $\{ G_{i} \mid i \in N_{q} \}$. The encoder is a transformer-based graph neural network (GNN) comprising an input projection layer and multiple TransformerConv~\cite{shi2020masked} blocks to model local and global vascular structures. Node embeddings are refined by linear layers with dropout and aggregated via global average pooling (GAP) to produce an embedding $\mathbf{g}_{q} \in \mathbb{R}^{c_{g}}$,
\begin{equation}
\mathbf{g}_{q} = \mathrm{GAP}(\mathcal{G}_{g}(G_{q})).
\end{equation}
For the sampled graph set, the individual embeddings are concatenated and passed through an multi-layer perceptron (MLP) $\mathcal{H}_{g}$ to yield a fused embedding $\mathbf{g}_{\{G_{i}\}} \in \mathbb{R}^{c_{g}}$:
\begin{equation} 
\mathbf{g}_{\{G_{i}\}} = \mathcal{H}_{g}(\mathrm{CONCAT}(\{\mathbf{g}_{i} \mid i \in N_{q} \})).
\end{equation}
Since both $\mathbf{g}_q$ and $\mathbf{g}_{\{G_i\}}$ are available during training, we average them to obtain the GPE $\mathbf{g}_{\text{train}} \in \mathbb{R}^{c_g}$, allowing both representations to participate in optimization while facilitating their alignment:
\begin{equation}
\mathbf{g}_{\text{train}} = \frac{1}{2} \left( \mathbf{g}_q + \mathbf{g}_{\{G_{i}\}}\right).
\end{equation}
During inference, since the graph $G_{q}$ cannot be constructed without the unavailable ground-truth vascular mask, the fused embedding $\mathbf{g}_{\{G_{i}\}}$ is directly used as the GPE.

\subsection{Reference-Guided Vascular-Prototype Learning}
Although RGPL enables robust GPE learning for capturing modality-aware structural priors, GPE alone has limited impact without direct interaction during upsampling, motivating the need for pixel-wise semantic enhancement. To address this limitation, we introduce vascular prototype embeddings (VPEs) for capturing modality-aware semantic prototypes of vessels and directly integrate them into the upsampling process of the mask decoder.

To accurately represent vessel features, VPEs are computed from ground-truth masks and multi-scale feature maps extracted by a parallel ResNet-style~\cite{he2016deep} CNN image encoder. Unlike vision transformers~\cite{dosovitskiy2020image} that operate at a uniform resolution, CNNs naturally learn hierarchical representations, making them well-suited for vascular prototype learning. Specifically, given a feature map $\mathbf{F}_{l} \in \mathbb{R}^{c_{l} \times h_{l} \times w_{l}}$ at scale $l$, the ground-truth mask is resized via interpolation to the corresponding resolution, yielding $\mathbf{M}_{l} \in \{ 0,1 \}^{h_{l} \times w_{l}}$. Masked average pooling is applied for extracting a scale-specific vascular embedding $\mathbf{p}_{l} \in \mathbb{R}^{c_{l}}$:
\begin{equation}
\mathbf{p}_{l} = \frac{\sum_{x,y} \mathbf{F}_{l}(x,y) \cdot \mathbf{M}_{l}(x,y)}{\sum_{x,y} \mathbf{M}_{l}(x,y)}.
\end{equation}
The multi-scale embeddings are concatenated and fused through an MLP to produce the prototype embedding $\mathbf{p}_{q} \in \mathbb{R}^{c_{p}}$ of the image.

Similarly to GPEs, VPEs should also be independent of ground-truth masks during inference while preserving robust modality-aware vessel semantic priors. Therefore, we adopt a similar strategy by introducing another graph encoder $\mathcal{G}_{p}$ and an MLP $\mathcal{H}_{p}$. Specifically, $\mathcal{G}_{p}$ first encodes each graph in $\{G_{i}\}$ individually to compute graph-level embeddings:
\begin{equation}
\mathbf{p}_{i} = \mathrm{GAP}(\mathcal{G}_{p}(G_{i})),
\end{equation}
which are then aggregated and passed through $\mathcal{H}_{p}$:
\begin{equation}
\mathbf{p}_{\{G_i\}} = \mathcal{H}_{p}\left(\mathrm{CONCAT}(\{\mathbf{p}_{i} \mid i \in N_{q}\})\right).
\end{equation}
During training, the VPE is obtained by averaging $\mathbf{p}_{q}$ and $\mathbf{p}_{\{G_{i}\}}$, still enabling both representations to be jointly optimized while encouraging their alignment:
\begin{equation}
\mathbf{p}_{\text{train}} = \frac{1}{2} (\mathbf{p}_{q} + \mathbf{p}_{\{G_{i}\}}),
\end{equation}
whereas $\mathbf{p}_{\{G_i\}}$ is directly used as the VPE during inference.

\subsection{Vessel-Aware Fine-Grained Decoding}
\label{sec:mask_decoder}
In the original SAM mask decoder, no skip connections are established between the image encoder and the mask decoder. Moreover, the mask decoder of SAM-B consists of only two upsampling stages followed by bilinear interpolation for resolution restoration, which limits the segmentation of fine-grained vascular structures. Therefore, we redesign the mask decoder by introducing skip connections from the CNN encoder, enabling direct integration of multi-scale local features. We further apply four progressive upsampling stages to directly reconstruct the segmentation mask at the original resolution.

Furthermore, we incorporate the learned VPE $\mathbf{p} \in \mathbb{R}^{c_{p}}$ for injecting modality-aware vessel semantic priors into the decoding process. Specifically, we introduce a vascular prototype attention (VPA) module at the end of each upsampling stage. Given a feature map $\mathbf{F}_{l} \in \mathbb{R}^{c_{l} \times h_{l} \times w_{l}}$ at scale $l$, the VPE $\mathbf{p}$ is first projected to the corresponding channel dimension $c_{l}$ through an MLP, yielding $\mathbf{p}_{l} \in \mathbb{R}^{c_{l}}$. A scaled dot-product similarity is then computed at each spatial location, followed by a sigmoid activation $\sigma$ for generating an attention map:
\begin{equation}
\mathbf{A}_{l} = \sigma \left( \frac{\mathbf{p}_{l}^{\top} \mathbf{F}_{l}}{\sqrt{c_{l}}} \right),
\end{equation}
which is then used to modulate the feature map via a residual gating mechanism:
\begin{equation}
\mathbf{F}_{l}^{'} = \mathbf{F}_{l} \cdot \left(1 + \mathbf{A}_{l}\right).
\end{equation}
This mechanism enables vessel-aware decoding for fine-grained segmentation of vascular structures.

\subsection{Loss Function}
The overall loss function consists of three components. Following\- the original design, the segmentation loss $\mathcal{L}_{seg}$ is defined as the sum of the Dice loss and binary cross-entropy (BCE) loss.

To facilitate the reference-guided representation learning schemes for GPEs and VPEs, we employ an $\mathcal{L}_{1}$ loss and a contrastive loss during training. The $\mathcal{L}_{1}$ loss directly minimizes the discrepancy between the estimated embeddings using the sampled graph set and the actual embeddings using the ground-truth masks, and the contrastive loss encourages modality-level semantic consistency by pulling embeddings from the same modality closer and pushing those from different modalities apart.
% The proposed reference-guided representation learning schemes aim to approximate the GPE $\mathbf{g}_{q}$ and VPE $\mathbf{p}_{q}$ of a query image $\mathbf{I}_{q}$ using the estimated embeddings $\mathbf{g}_{\{G_{i}\}}$ and $\mathbf{p}_{\{G_{i}\}}$ from the retrieved graph set $\{G_{i}\}$. To this end, we employ $\mathcal{L}_{1}$ loss and a contrastive loss $\mathcal{L}_{con}$~\cite{radford2021learning} to enforce GPE alignment:
Accordingly, the GPE alignment loss is defined as
\begin{equation}
\mathcal{L}_{g} = \lambda_{ag} \mathcal{L}_{1}(\mathbf{g}_{q}, \mathbf{g}_{\{G_{i}\}}) + \lambda_{cg} \mathcal{L}_{con}(\mathbf{g}_{q},\mathbf{g}_{\{G_{i}\}}).
\end{equation}
Similarly, the VPE alignment loss is defined as
\begin{equation}
\mathcal{L}_{p} = \lambda_{ap} \mathcal{L}_{1}(\mathbf{p}_{q}, \mathbf{p}_{\{G_{i}\}}) + \lambda_{cp} \mathcal{L}_{con}(\mathbf{p}_{q},\mathbf{p}_{\{G_{i}\}}).
\end{equation}
Correspondingly, the overall loss function is formulated as:
\begin{equation}
\mathcal{L} = \mathcal{L}_{seg} + \mathcal{L}_{g} 
+ \mathcal{L}_{p}.
\end{equation}

\section{Experiments}
\subsection{Datasets}
We collected 19 public datasets spanning six imaging modalities to construct a comprehensive benchmark. Dataset details are summarized in Table~\ref{table1}. Except for ORVS~\cite{sarhan2021transfer}, all datasets were merged and split into training and test sets at the 80\% to 20\% ratio, yielding 3349 and 844 samples, respectively. The ORVS dataset, comprising 49 samples, was held out as a zero-shot test set. All images were converted to 3~channels and resized to $1024 \times 1024$ pixels using bicubic interpolation. Pixel values were normalized to the range $\left [ 0,1 \right ]$. During training, we applied data augmentation, including random contrast, gamma correction, brightness adjustment, and Gaussian noise.

\begin{table}[t]
\centering
\fontsize{9}{12}\selectfont
\setlength{\tabcolsep}{1mm}
\begin{tabular}{lc}
\toprule
Dataset & Modality \\
\midrule
FS-CAD~\cite{zeng2024pretrained} & X-Ray \\
XACV~\cite{wu2025denver} & X-Ray \\
XCAD~\cite{ma2021self} & X-Ray \\
Dr-SAM~\cite{zohranyan2024dr} & X-Ray \\
DIAS~\cite{liu2024dias} & DSA \\
DSCA~\cite{zhang2025dsca} & DSA \\
FPD~\cite{bano2020deep} & PF \\
OCTA-500 3mm~\cite{li2024octa} & OCTA \\
OCTA-500 6mm~\cite{li2024octa} & OCTA\\
ROSE-1~\cite{ma2020rose} & OCTA \\
ROSSA~\cite{ning2024accurate} & OCTA \\
RAVIR~\cite{hatamizadeh2022ravir,hatamizadeh2020artificial} & SLO \\
ARIA~\cite{farnell2008enhancement} & Fundus \\
CHASE-AV~\cite{fraz2012ensemble} & Fundus \\
DRIVE~\cite{staal2004ridge} & Fundus \\
FIVES~\cite{jin2022fives} & Fundus\\
LES-AV~\cite{orlando2018towards} & Fundus \\
STARE~\cite{hoover2000locating} & Fundus \\
\hline
ORVS~\cite{sarhan2021transfer} & Fundus \\
\bottomrule
\end{tabular}
\caption{A summary of the datasets. DSA: digital subtraction angiography, PF: placental fetoscopy, OCTA: optical coherence tomography angiography, SLO: scanning laser ophthalmoscopy.}
\label{table1}
\end{table}

\subsection{Implementation Details}
Given the limited data volume, rather than fully fine-tuning the SAM image encoder, we used adapter modules~\cite{chen2022adaptformer} for each transformer block of the image encoder to efficiently adapt to vascular images. During training, the original encoder weights were frozen, and only the adapter parameters were optimized. To initialize ReG-SAM, we used the pretrained MedSAM-B weights~\cite{ma2024efficient}.

ReG-SAM was implemented in PyTorch 2.6.0 and trained on an NVIDIA A100 GPU (80 GB memory) for 25 epochs using the AdamW optimizer with an initial learning rate of $1\times10^{-4}$ and a batch size of 2. A reduction-on-plateau scheduler was employed to decay the learning rate. The loss coefficients $\lambda_{ag}, \lambda_{cg}, \lambda_{ap}, \lambda_{cp}$ are set to 0.05, 0.05, 0.1, 0.1, respectively. Since the segmentation loss remains the primary objective, these coefficients mainly serve as regularization terms. Notably, the vascular prototype embedding, which emphasizes local fine-grained processing, is more closely aligned with the nature of vessel segmentation and was therefore assigned higher weights. The value of $k$ in database sampling was set to 3 through extensive experiments.

\subsection{Compared Methods}
We selected five representative benchmark methods for comparison. SAM-Med2D~\cite{cheng2023sam} and MedSAM~\cite{ma2024segment} are widely used foundation models in medical image segmentation. SAM-HQ~\cite{ke2023segment}, designed for high-fidelity natural image segmentation, was adopted here as a strong baseline. ASPS~\cite{li2024asps}, a strong polyp segmentor, is included due to its potential in vessel segmentation. OVS-Net~\cite{song2025optimized} is the current state-of-the-art general 2D vessel segmentor. Regarding prompting, SAM-Med2D and SAM-HQ support both bounding-box and point prompts, so we evaluated two variants for each. The prompting for MedSAM and OVS-Net follows their original designs, emphasizing bounding-box and point prompts, respectively, while ASPS is prompt-free. For a fair evaluation, all the methods were retrained on the benchmark dataset. Notably, given the large domain gap between natural and vascular images, we initialized SAM-HQ with MedSAM weights. For quantitative evaluation, we adopted the Dice coefficient (Dice) and intersection-over-union (IoU) as metrics to evaluate the segmented vascular morphology.

\subsection{Quantitative Results}
\begin{figure}[!t]
\centering
\includegraphics[width=\columnwidth]{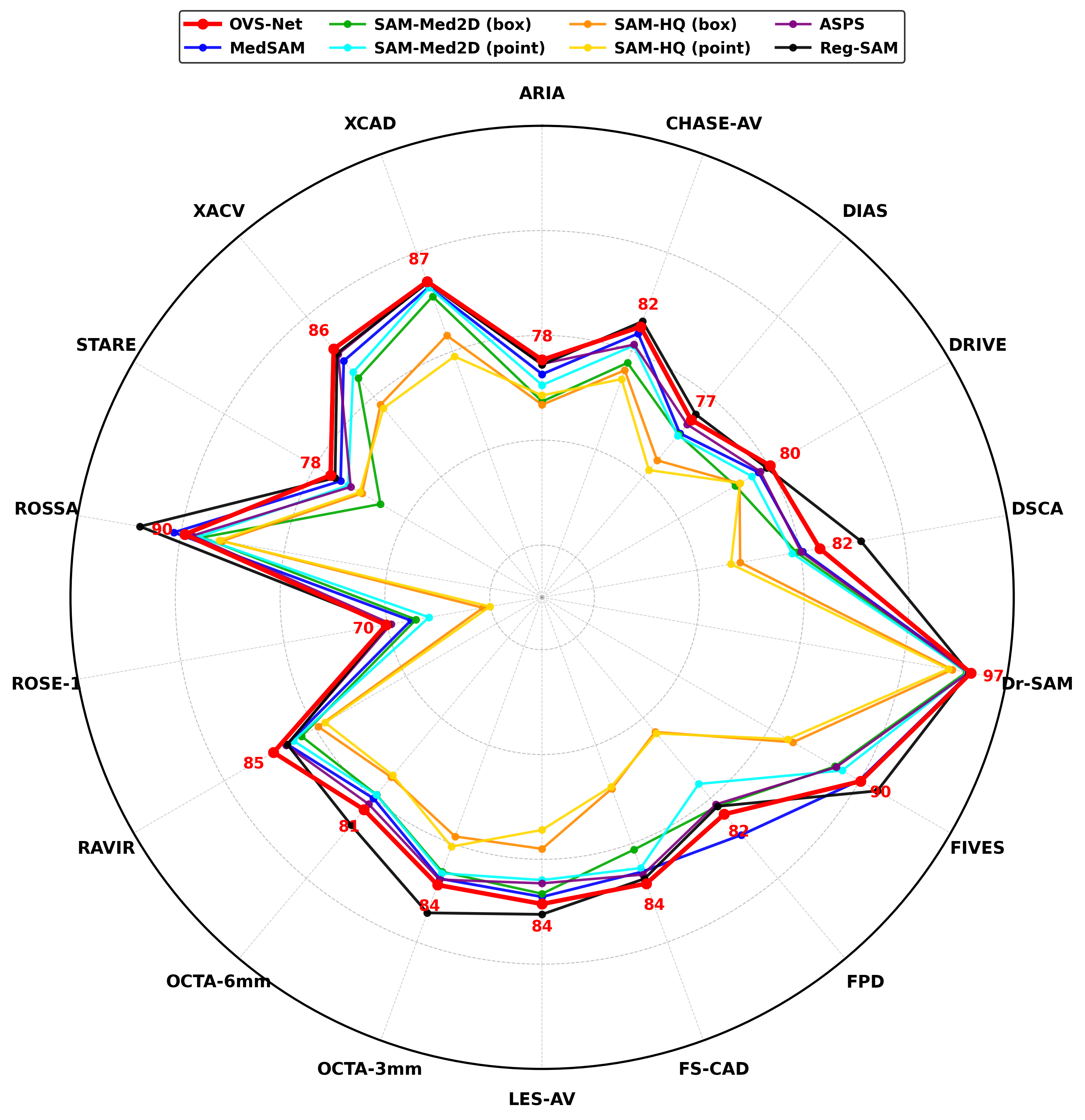}
\caption{Radar chart comparing dataset-wise Dice scores of different methods.}
\label{fig2}
\end{figure}

Quantitative results reported in  Table~\ref{table2}, grouped by modality, demonstrate that ReG-SAM achieves the best overall segmentation performance on both regular and zero-shot test sets. Although it does not rank first in each modality, it consistently excels on the three most challenging modalities, including Fundus, DSA, and OCTA, where complex vascular morphologies with numerous thin and branching vessels make accurate segmentation particularly difficult. On the zero-shot set, ReG-SAM also outperforms OVS-Net, demonstrating strong generalization ability. Notably, these improvements are achieved without requiring any prompt inputs, further highlighting the practicality of ReG-SAM. Figure~\ref{fig2} provides a dataset-level comparison, showing that ReG-SAM delivers more balanced performance by achieving the best results on most datasets while exhibiting only limited performance degradation on a few, consistent with the modality-wise analysis.

\begin{table*}[t]
\centering
\fontsize{9}{12}\selectfont
\setlength{\tabcolsep}{1mm}

\begin{tabular}{lcc|cc|cc|cc}
\toprule
\multirow{2}{*}{Method} 
& \multicolumn{2}{c}{Fundus} 
& \multicolumn{2}{c}{DSA} 
& \multicolumn{2}{c}{X-Ray} 
& \multicolumn{2}{c}{PF} \\
\cmidrule(lr){2-3} \cmidrule(lr){4-5} \cmidrule(lr){6-7} \cmidrule(lr){8-9}
& Dice$\uparrow$ & IoU$\uparrow$
& Dice$\uparrow$ & IoU$\uparrow$
& Dice$\uparrow$ & IoU$\uparrow$
& Dice$\uparrow$ & IoU$\uparrow$ \\
\midrule

ASPS (auto) & 85.17$\pm$7.72 & 74.83$\pm$9.87 & 79.34$\pm$4.75 & 66.00$\pm$6.45 & 91.51$\pm$5.90 & 84.89$\pm$9.78 & 80.79$\pm$11.35 & 69.12$\pm$14.22 \\
MedSAM (box) & 87.04$\pm$8.54 & 77.89$\pm$11.34 & 79.22$\pm$4.85 & 65.86$\pm$6.57 & 91.28$\pm$6.28 & 84.56$\pm$10.33 & \textbf{84.59$\pm$8.14} & \textbf{74.08$\pm$11.29} \\
OVS-Net (point)  & \textit{87.44$\pm$8.19} & \textit{78.46$\pm$10.87} & \textit{80.89$\pm$4.57} & \textit{68.15$\pm$6.33} & \textbf{91.86$\pm$5.75} & \textbf{85.46$\pm$9.59} & \textit{82.02$\pm$10.66} & \textit{70.75$\pm$13.70} \\
SAM-HQ (box) & 80.80$\pm$8.58 & 68.51$\pm$10.29 & 73.74$\pm$5.29 & 58.69$\pm$6.60 & 87.82$\pm$8.75 & 79.32$\pm$13.33 & 71.77$\pm$18.08 & 58.61$\pm$19.06 \\
SAM-HQ (point) & 80.41$\pm$8.08 & 67.89$\pm$9.65 & 72.78$\pm$5.54 & 57.51$\pm$6.79 & 87.18$\pm$9.07 & 78.38$\pm$13.75 & 71.95$\pm$18.80 & 58.97$\pm$19.34 \\
SAM-Med2D (box) & 84.38$\pm$8.80 & 73.84$\pm$11.36 & 78.74$\pm$5.03 & 65.21$\pm$6.69 & 90.31$\pm$7.19 & 83.09$\pm$11.52 & 81.04$\pm$10.33 & 69.28$\pm$13.38 \\
SAM-Med2D (point) & 85.35$\pm$8.45 & 75.24$\pm$10.97 & 78.37$\pm$5.33 & 64.74$\pm$6.98 & 90.79$\pm$6.72 & 83.80$\pm$10.89 & 78.24$\pm$11.84 & 65.70$\pm$14.86 \\
ReG-SAM (auto) & \textbf{88.78$\pm$8.93} & \textbf{80.79$\pm$12.30} & \textbf{84.19$\pm$5.08} & \textbf{73.02$\pm$7.37} & \textit{91.56$\pm$5.84} & \textit{84.96$\pm$9.70} & 81.03$\pm$10.95 & 69.36$\pm$13.76 \\

\midrule

\multirow{2}{*}{Method} 
& \multicolumn{2}{c}{OCTA} 
& \multicolumn{2}{c}{SLO} 
& \multicolumn{2}{c}{Average} 
& \multicolumn{2}{c}{Zero Shot} \\
\cmidrule(lr){2-3} \cmidrule(lr){4-5} \cmidrule(lr){6-7} \cmidrule(lr){8-9}
& Dice$\uparrow$ & IoU$\uparrow$
& Dice$\uparrow$ & IoU$\uparrow$
& Dice$\uparrow$ & IoU$\uparrow$
& Dice$\uparrow$ & IoU$\uparrow$ \\
\midrule

ASPS (auto) & 85.97$\pm$5.15 & 75.72$\pm$7.45 & \textit{83.20$\pm$1.88} & \textit{71.28$\pm$2.78} & 85.90$\pm$7.80 & 76.02$\pm$10.95 & 64.00$\pm$5.97 & 47.34$\pm$6.40 \\
MedSAM (box) & \textit{86.89$\pm$6.19} & \textit{77.31$\pm$9.08} & 83.06$\pm$1.85 & 71.07$\pm$2.74 & \textit{87.06$\pm$7.62} & \textit{77.83$\pm$11.08} & 64.77$\pm$5.41 & 48.13$\pm$5.88 \\
OVS-Net (point)  & 86.66$\pm$5.18 & 76.81$\pm$7.56 & \textbf{84.59$\pm$1.85} & \textbf{73.34$\pm$2.79} & 87.04$\pm$7.63 & 77.78$\pm$10.89 & \textit{65.27$\pm$5.67} & \textit{48.71$\pm$6.16} \\
SAM-HQ (box) & 82.62$\pm$6.04 & 70.80$\pm$8.09 & 79.67$\pm$3.63 & 66.36$\pm$4.96 & 81.41$\pm$10.54 & 69.81$\pm$13.24 & 62.08$\pm$5.73 & 45.26$\pm$5.94 \\
SAM-HQ (point) & 82.86$\pm$6.21 & 71.17$\pm$8.26 & 78.91$\pm$3.94 & 65.34$\pm$5.31 & 81.21$\pm$10.66 & 69.53$\pm$13.22 & 61.03$\pm$5.87 & 44.17$\pm$5.99 \\
SAM-Med2D (box) & 84.96$\pm$5.36 & 74.21$\pm$7.54 & 81.51$\pm$2.80 & 68.88$\pm$3.97 & 85.07$\pm$8.07 & 74.81$\pm$11.37 & 64.24$\pm$5.30 & 47.54$\pm$5.69 \\
SAM-Med2D (point) & 85.15$\pm$5.63 & 74.53$\pm$7.89 & 82.38$\pm$2.17 & 70.10$\pm$3.14 & 85.14$\pm$8.49 & 74.99$\pm$11.83 & 63.86$\pm$5.73 & 47.17$\pm$6.03 \\
ReG-SAM (auto) & \textbf{90.18$\pm$6.40} & \textbf{82.69$\pm$9.86} & 83.07$\pm$1.68 & 71.08$\pm$2.50 & \textbf{88.63$\pm$8.24} & \textbf{80.45$\pm$11.91} & \textbf{66.20$\pm$5.67} & \textbf{49.74$\pm$6.25} \\

\bottomrule
\end{tabular}
\caption{Modality-wise, average, and zero-shot segmentation performance. The best results are highlighted in \textbf{bold} and the second-best in \textit{italic}.}
\label{table2}
\end{table*}

\subsection{Qualitative Results}
\begin{figure*}[!t]
\centering
\includegraphics[width=\textwidth]{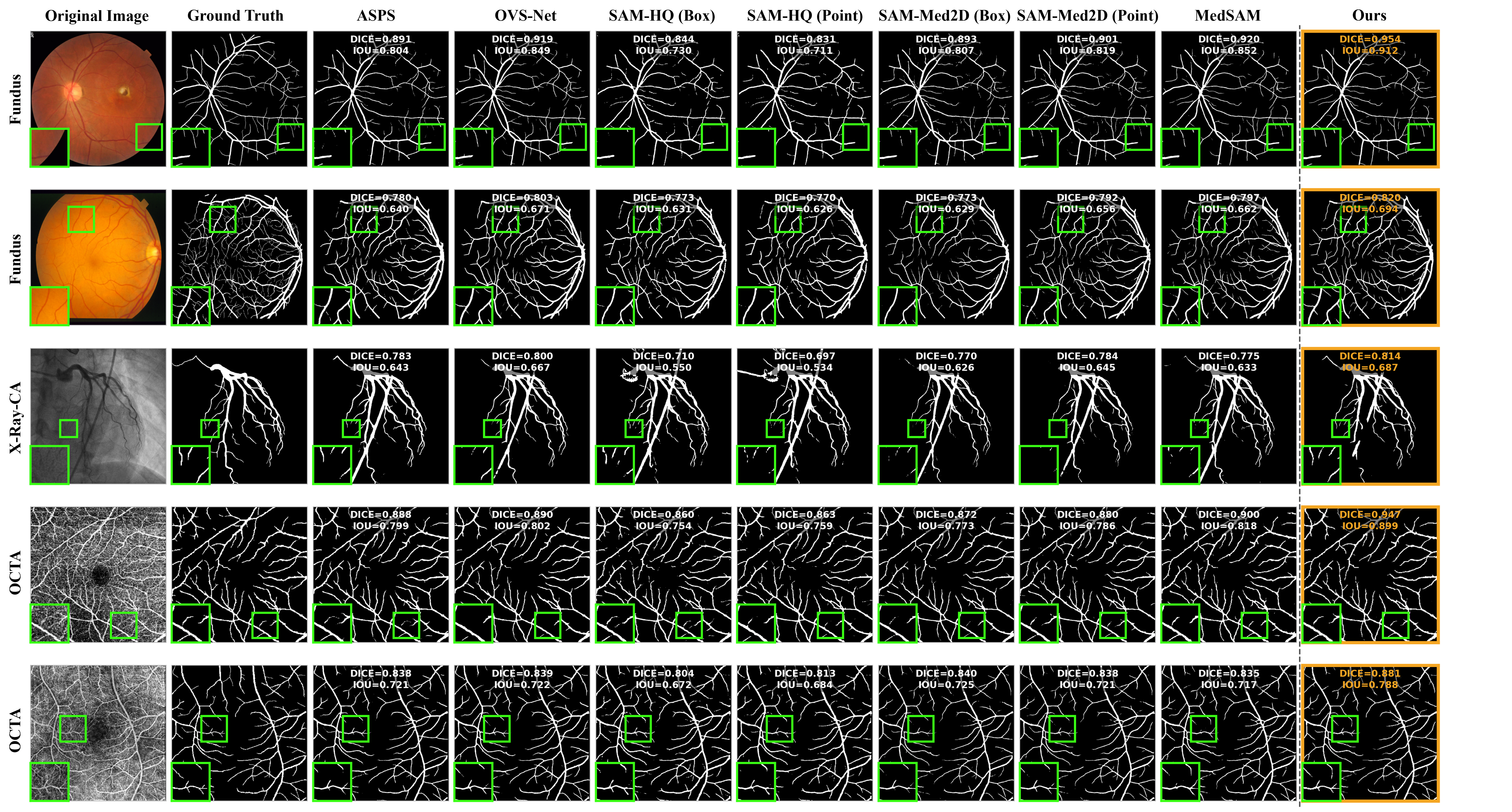}
\caption{Qualitative comparison of different methods. Challenging regions are highlighted by green boxes and enlarged in at the bottom left corner. Evaluation metrics for each case are shown at the top, with the highest values highlighted in yellow.}
\label{fig3}
\end{figure*}

Figure~\ref{fig3} presents qualitative results to provide a more comprehensive evaluation of ReG-SAM. We highlight several challenging vascular modalities characterized by complex, highly branched vessel structures. As shown in the regions marked by green boxes, baseline methods tend to under-segment fine vessels, whereas ReG-SAM preserves these fine structures more effectively. Although ReG-SAM still misses some vessel details, it consistently demonstrates a stronger capability to capture intricate vascular morphology, resulting in more accurate segmentation. These qualitative results further validate the effectiveness of ReG-SAM for fine-grained vessel segmentation.

\subsection{Ablation Studies}
Ablation studies are conducted to evaluate the key components of ReG-SAM, with results summarized in Table~\ref{table3}. The first three rows examine the architecture by reducing the decoder upsampling stages from four to the original two in SAM, or by removing either GPEs or VPEs. Notably, since VPEs are designed to work jointly with the VPA module during upsampling, both VPE and VPA are removed in this ablation. Reducing the number of upsampling stages consistently degrades performance, demonstrating the importance of gradual, learnable upsampling for recovering complex vascular structures. Removing either GPEs or VPEs also degrades performance, with a larger drop observed after removing VPEs, highlighting the importance of vessel-aware fine-grained decoding. Employed in a pixel-wise manner, VPEs and VPA provide rich local semantic guidance for vessel segmentation, whereas GPEs capture global structural priors that provide coarse but informative morphological context.

The next two rows evaluate the representation loss terms,  $\mathcal{L}_{g}$ and $\mathcal{L}_{p}$, for GPEs and VPEs, respectively. As shown in Table\-~\ref{table3}, removing either loss degrades performance, indicating that both losses effectively regularize the learned representations by encouraging modality-aware embedding alignment during training. Consequently, more discriminative structural and semantic representations can be obtained during inference, leading to improved segmentation accuracy\-.

\begin{table}[t]
\centering
\begin{tabular}{lcc}
\toprule
\multirow{2}{*}{Configurations} 
& \multicolumn{2}{c}{Regular} \\
\cmidrule(lr){2-3}
& Dice$\uparrow$ & IoU$\uparrow$ \\
\midrule
w/o 4 Upsampling & 88.08$\pm$8.46 & 79.61$\pm$12.23 \\
w/o GPEs & 88.34$\pm$8.32 & 80.00$\pm$11.99 \\
w/o VPEs \& VPA & 85.10$\pm$10.36 & 75.32$\pm$14.05 \\
w/o $\mathcal{L}_{g}$ & 88.52$\pm$8.24 & 80.28$\pm$11.98 \\
w/o $\mathcal{L}_{p}$ & 86.92$\pm$8.09 & 77.67$\pm$11.46 \\
ReG-SAM & \textbf{88.63$\pm$8.24} & \textbf{80.45$\pm$11.91} \\
\toprule
\multirow{2}{*}{Configurations} 
& \multicolumn{2}{c}{Zero Shot} \\
\cmidrule(lr){2-3}
& Dice$\uparrow$ & IoU$\uparrow$ \\
\midrule
w/o 4 Upsampling & 64.91$\pm$5.67 & 48.30$\pm$6.14 \\
w/o GPEs & 64.27$\pm$5.64 & 47.60$\pm$6.03 \\
w/o VPEs \& VPA & 62.97$\pm$5.58 & 46.19$\pm$5.88 \\
w/o $\mathcal{L}_{g}$ & 65.46$\pm$5.78 & 48.93$\pm$6.32 \\
w/o $\mathcal{L}_{p}$ & 61.57$\pm$5.51 & 44.71$\pm$5.72 \\
ReG-SAM & \textbf{66.20$\pm$5.67} & \textbf{49.74$\pm$6.25} \\
\bottomrule
\end{tabular}
\caption{Experimental results of the ablation studies on architectures and loss functions. The best results are highlighted in \textbf{bold}.}
\label{table3}
\end{table}

\subsection{Comparison on Sampling Strategies}
\begin{figure}[!t]
\centering
\includegraphics[width=\columnwidth]{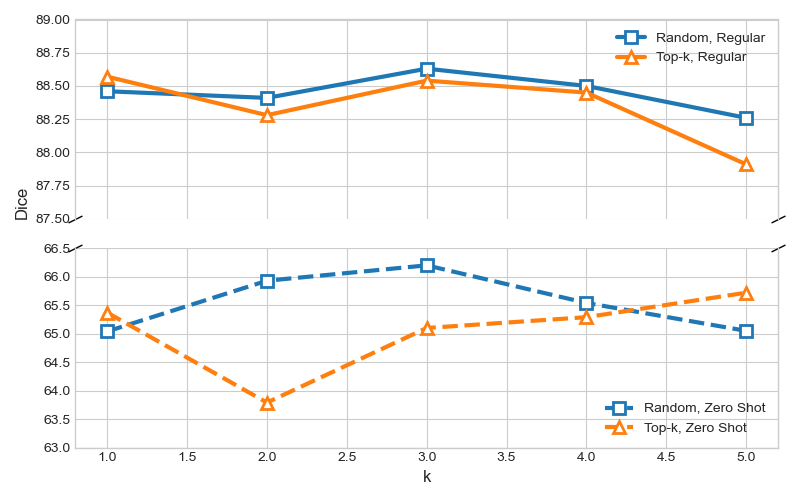}
\caption{Performance comparison on the regular and zero-shot test sets using random and top-$k$ sampling strategies.}
\label{fig4}
\end{figure}

Since our goal is to learn robust, modality-aware vascular priors, we adopted random sampling instead of similarity-based top-$k$ sampling to select reference graphs from the vascular database. For top-$k$ sampling, image embeddings were extracted using DINOv2~\cite{oquab2023dinov2}, and the $k$ most similar images were retrieved based on cosine similarity to construct the graph set. Results of comparing both strategies with $k$ ranging from 1 to 5 are shown in Figure~\ref{fig4}.

Overall, random sampling outperforms top-$k$ sampling under the current dataset setting. Random sampling with $k=3$ achieves the best performance in both cases. Smaller $k$ values provide insufficient information for effective alignment, while larger values increase fitting difficulty and introduce more noise, leading to performance degradation. In addition, random sampling is more stable across different $k$ values, likely because it exposes the model to more diverse graph combinations. In contrast, top-$k$ sampling relies on a limited set of neighbors, resulting in poorer generalization to unseen graph sets, especially under limited data conditions.

% As a result, models trained with top-$k$ sampling generalize worse to unseen graph sets, particularly under limited data conditions where increased sampling diversity is crucial for generalization.

\section{Discussion and Conclusion}
In this paper, we propose ReG-SAM, a reference graph-guided foundation model for 2D vessel segmentation built upon SAM. Motivated by recent advances in general vessel segmentation, we introduce two embedding representations, GPEs and VPEs, to enhance SAM framework for vessel segmentation. To eliminate their reliance on ground-truth masks during inference while promoting learning robust modality-aware representations, we construct a vascular database and introduce two reference graph-guided representation learning schemes to estimate GPEs and VPEs from randomly sampled graph set. Comprehensive evaluations on 19 public vascular datasets demonstrate that ReG-SAM consistently outperforms existing methods, achieving more accurate overall segmentation and stronger zero-shot generalization.

However, the performance of ReG-SAM is still limited by several factors. First, vessel annotation is highly challenging and labor-intensive, resulting in substantially fewer available samples compared to tasks such as organ segmentation. Even after collecting multiple public datasets, the total data volume remains below 5000 samples, limiting the full potential of foundation models. This is particularly relevant to our framework, as its reference-based mechanism could benefit from a larger and more diverse database, potentially enabling more effective reference selection and alternative sampling strategies. Furthermore, highly accurate predictions may be mistaken for over-segmentation due to low-quality ground-truth annotations, leading to underestimated performance. We therefore expect that larger, higher-quality, and more diverse vessel datasets will further improve ReG-SAM and widen its performance gap over existing methods.

\appendix

% \section*{Acknowledgments}
% AAAI is especially grateful to Peter Patel Schneider for his work in implementing the original aaai.sty file, liberally using the ideas of other style hackers, including Barbara Beeton. We also acknowledge with thanks the work of George Ferguson for his guide to using the style and BibTeX files --- which has been incorporated into this document --- and Hans Guesgen, who provided several timely modifications, as well as the many others who have, from time to time, sent in suggestions on improvements to the AAAI style. We are especially grateful to Francisco Cruz, Marc Pujol-Gonzalez, and Mico Loretan for the improvements to the Bib\TeX{} and \LaTeX{} files made in 2020.

% The preparation of the \LaTeX{} and Bib\TeX{} files that implement these instructions was supported by Schlumberger Palo Alto Research, AT\&T Bell Laboratories, Morgan Kaufmann Publishers, The Live Oak Press, LLC, and AAAI Press. Bibliography style changes were added by Sunil Issar. \verb+\+pubnote was added by J. Scott Penberthy. George Ferguson added support for printing the AAAI copyright slug. Additional changes to aaai2027.sty and aaai2027.bst have been made by Francisco Cruz, Marc Pujol-Gonzalez, and Mico Loretan.

% \bigskip
% \noindent Thank you for reading these instructions carefully. We look forward to receiving your electronic files!

\fontsize{9.0pt}{10.0pt} \selectfont
\bibliography{aaai2027}

% Check whether the conference requires a reproducibility checklist to be included in the paper.
% If so, you can uncomment the following line and ajust the path to include it.
% \input{ReproducibilityChecklist.tex}

\end{document}